\documentclass[letterpaper, 10 pt, conference]{ieeeconf}  

\IEEEoverridecommandlockouts                              

\title{\LARGE \bf
Next-Best-View Estimation based on Deep Reinforcement Learning for Active Object Classification
}

\usepackage{hyperref}
\usepackage{url}
\usepackage{graphicx}
\usepackage{float}
\usepackage{subfig} 
\usepackage{multirow}
\author{Christian Korbach$^{1}$, Markus D. Solbach$^{2}$, Raphael Memmesheimer$^{1}$, Dietrich Paulus$^{1}$, John K. Tsotsos$^{2}$
	\thanks{$^{1}$Active Vision Group, University of Koblenz, Germany.
		\mbox{{\tt\small \{ckorbach, raphael, paulus\}@uni-koblenz.de}}}
	\thanks{$^{2}$Department of Electrical Engineering and Computer Science, York University, Canada.
		\mbox{{\tt\small\{solbach, tsotsos\}@eecs.yorku.ca}}}
}

\begin{document}

\maketitle
\thispagestyle{empty}
\pagestyle{empty}


\begin{abstract}
The presentation and analysis of image data from a single viewpoint are often not sufficient to solve a task. 
Several viewpoints are necessary to obtain more information. 
The \textit{next-best-view} problem attempts to find the optimal viewpoint with the greatest information gain for the underlying task.
In this work, a robot arm holds an object in its end-effector and searches for a sequence of next-best-view to explicitly identify the object.
We use Soft Actor-Critic (SAC), a method of deep reinforcement learning, to learn these next-best-views for a specific set of objects.
The evaluation shows that an agent can learn to determine an object pose to which the robot arm should move an object.
This leads to a viewpoint that provides a more accurate prediction to distinguish such an object from other objects better.
We make the code publicly available for the scientific community and for reproducibility.\footnote[3]{\url{https://github.com/ckorbach/nbv_rl}}
\end{abstract}


\section{Introduction}
\label{sec:introduction}
\begin{figure}[H]
	\centering$
	\begin{array}{cc}
	\includegraphics[width=.3\linewidth]{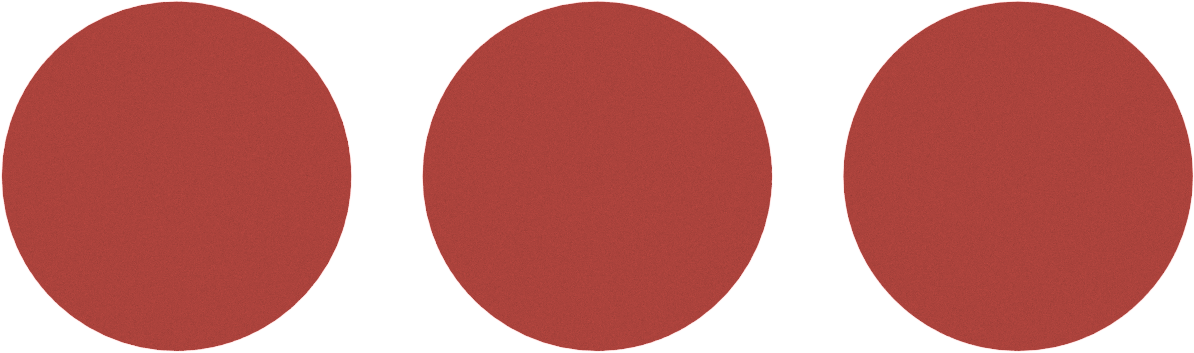} 
	\\
	\includegraphics[width=.3\linewidth]{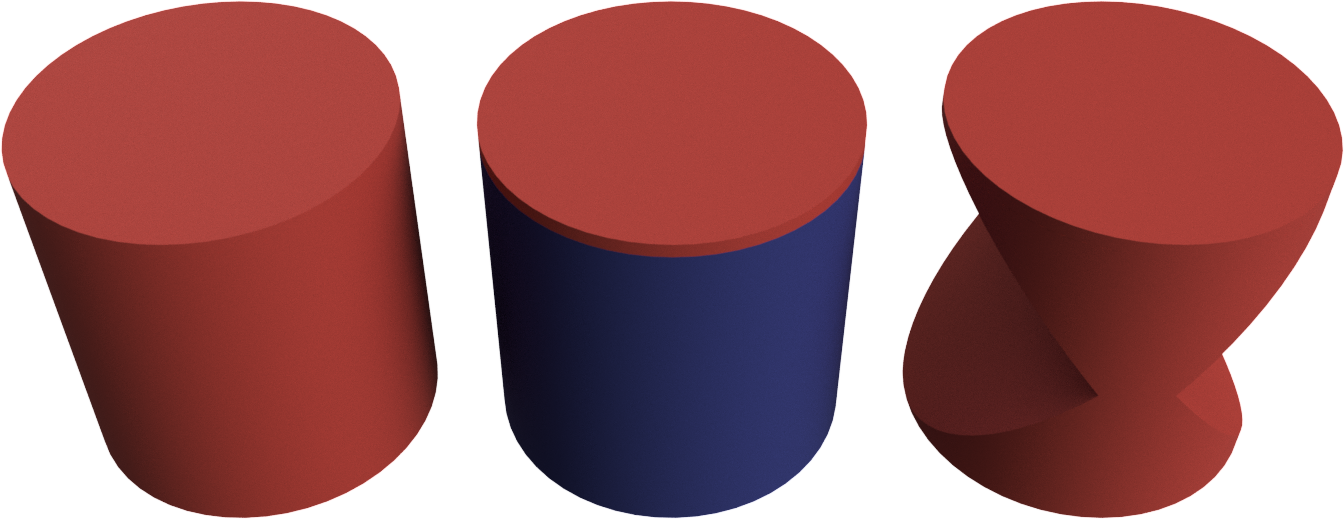}\\
	\includegraphics[width=.3\linewidth]{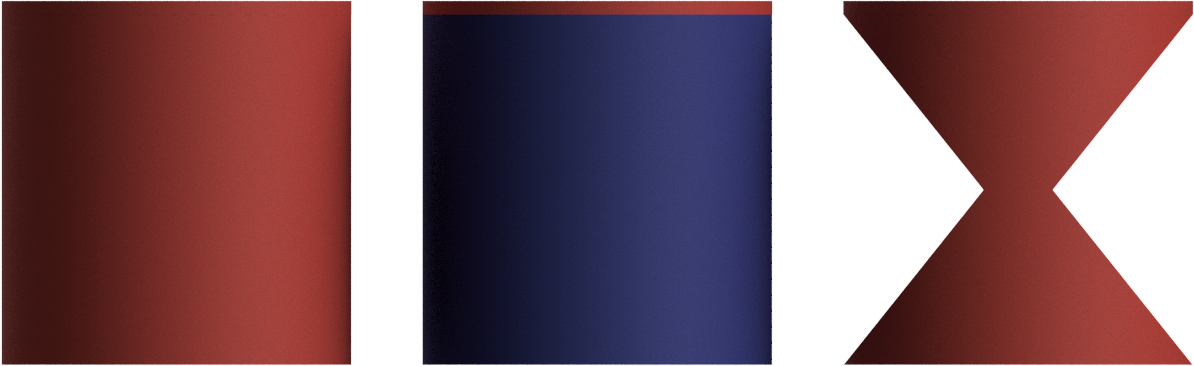}
	\end{array}$
	\caption{Motivational example for next-best-view estimation. In the top view all objects look same and no clear object class can be associated. Changing the camera view allows to observe more specific object features. The top and bottom rows of image projections are know as \textit{degenerate views} \cite{DIC99degeneracy}.}
	\label{fig:example_motivation}
\end{figure}

Humans have the ability to actively observe objects in order to classify them efficiently and reliably. It is an omnipresent ability that is used continuously and effortlessly. 
Computer-based object classification is already present in a wide range of applications, such as optical character classification, industrial inspection systems, medical imaging, biometrics, and others. 
However, the scientific challenge is that a single object can generate an infinite number of 2D images depending on the object position, illumination, and background, thus significantly changing the appearance of the object  \cite{szeliski2010computer} and leading to system failure.
Furthermore, \cite{Tsotsos1989, Bajcsy2018} show that visual object classification, in general, is a hard problem. See Fig. \ref{fig:example_motivation} as an example how three different viewpoints can change the object appearance notably.
\\
Modern object classification systems are mostly passive – acting as a signal-processing procedure without control over the signal. 
Visual perception is a problem of control of data acquisition and not necessarily one of signal processing  \cite{Bajcsy1985,Bajcsy1988}.
It is argued that the activity of perception is exploratory, probing, searching.
The analogy used is that ``percepts do not simply fall onto sensors as rain falls onto the ground. We do not just see, we look.''
This becomes even more clear when we think of a toolbox filled with many screwdrivers in which we want to find the right one for our task.
We must change viewpoints to classify the correct tool. 
Enabling a vision system with the ability to choose the next observation, complexities of vision tasks can be simplified \cite{Andreopoulos2013a}, such as visual object classification.
In this paper, we propose an approach for estimating a sequence of next-best-view (NBV) for visual object classification. 
The goal is to use image interpretations to control the viewpoint of the object under investigation purposefully. 
To accomplish this, we follow the widely used two-part approach for active object classification \cite{Andreopoulos2013a}:

\begin{enumerate}
    \item Object Classifier,
    \item Next-Best-View Estimator.
\end{enumerate}

Research activity and improvements in both fields are not at par. 
Tremendous improvements have been seen in the field of object classification, especially with the rise of a machine learning method called deep learning. 
\\We, therefore, focus on the second part, the next-best-view estimator.
We use a deep reinforcement learning method called \textit{Soft Actor-Critic} (SAC) by \cite{HAAR18sacaa} to learn robot arm movements that will gain the maximum additional information about the object under investigation.
The basic idea of reinforcement learning is to capture the most critical aspects of the real problem facing a learning agent interacting with its environment to achieve a goal \cite{Sutton1998}.
In our case, we use the object classifier's output as the measurement and reward computation of the NBV estimator.
A simulation environment is used to improve the learning process to achieve higher sample efficiency and less time expenditure.
The simulation is based on a real-world setup to provide a fundamental basis for transfer to real applications.
\\
The contribution of this paper is as follows:
First, we will present our experimental setup, outline the structure of our approach and explain the difficulties of self-occlusion in context of this work.
Then, we introduce the used dataset in more detail and present the results of the object classifier and the next-best-view estimator.
Finally, we show the results and give a conclusion.

\section{Related Work}
\label{sec:related_work}
In this work, we use a selection of objects from the \textit{TEOS (The Effect of Self-Occlusion)} dataset by Solbach and Tsotsos \cite{solbach2021blocks}, inspired by Shepard and Metzler (1971) \cite{SHE71ds}.
They are designed to investigate the effect of self-occlusion on image classifiers.
The objects are made of different degrees of complexity, corresponding to the number of components of an object, share a common coordinate system and distinguished only by their geometry.
They are also used to study a 3D version of the well-known same-different task with active observer \cite{CAR93_samediff}.
\\
Geometric solutions based on 2D sensors are common methods to solve the next-best-view problem.
They often tried to find features and track them to obtain a model.
The problem was often broken down into a ``begin-game``, ``mid-game``, and  ``end-game`` \cite{OHH98}.
First, the overall object shape was obtained by a quick scan, and then more large unseen areas would be scanned.
In the end, the most difficult concavities were explored.
However, solutions were mostly proposed for individual stages rather than all together.
Approaches of the begin- and mid-game often scan all surfaces by making and combining multiple images from different viewpoints \cite{BANTA95}\cite{PITO95range}\cite{P96}.
However, it is remarkable that almost no one has tried to solve the endgame \cite{MB93}.
This is not that easy, since objects are often non-convex and consist of self-occlusion.
Some approaches use multi-sensor systems, while others move only one sensor to different viewpoints.
Wilkes and Tsotsos (1992) \cite{WIL92cvpr} present a system in which a 2D camera mounted on a robotic arm moves around an object resting on a table.
They identify a set of lines as feature vectors and calculate a new point of view based on the knowledge gained about the object.
The camera moves to the new position to identify a new set of feature vectors and compares them with the current data using an indexing scheme until the object is identified.
This sequence of actions is a universal application strategy of active object recognition approaches.
\\
In the last decade, modern approaches mostly used deep learning.
With the further rise of depth sensors in cameras, active object recognition research has been extended to 3-dimensional space.
The observed target objects are mostly represented by polygon meshes and voxel grids and often imported or exported as CAD models.
To avoid using a predefined set of viewpoints, we again search for viewpoints that provide the most additional information until we reach the goal of the application. 
The field includes a wide range of applications, such as reconstruction, pose estimation and classification of 3D objects.
Some methods aim to reconstruct unknown objects to be able to create whole datasets autonomously \cite{MEN20reconstruction}\cite{DAUD17reconstruction}.
Other algorithms use next-best-view estimation to compute and recover 6D object poses \cite{DOUM15objectpose_nbv} \cite{SOCK19pose_drl}\cite{GEO18}.
Doumanoglou et al. (2015) \cite{DOUM15objectpose_nbv} select the viewpoint by entropy reduction, where entropy is the uncertainty of the detection in the new viewpoint.
Classification methods are mostly based on CNN's and extend them by adding a view pooling layer.
The networks are trained on multiple views of objects, e.g. through multiple images from different viewpoints \cite{KAN16rotationnet} or CAD models \cite{SU15multi-view-shape}\cite{WU14shapenet}.
\\
Krainin et al. (2011) \cite{KRAIN11nbv_completion} propose a method with a static sensor and an object that is dynamically moved around by a robot arm.
They compute next-best-views to find optimal gripping points to grasp and move the object to reconstruct it as a 3D model.
\\
In addition to Deep Learning, Deep Reinforcement Learning (DRL) methods are widely used in robotics to train robots how to move to solve a task.
Sock et al. (2019) \cite{SOCK19pose_drl} use DRL to train an agent that decides which next viewpoint is the most promising to estimate the object's pose.
Peralta et al. \cite{scanRL} propose an NBV policy that scans houses with fewer number of steps compared to their baseline. 
Schmid et al. (2019) \cite{schmid2019explore} train an agent to explore the environment to find a target object.
\\
Past work provides approaches to estimate the NBV for Active Object Reconstruction and Active Visual Object Search using mobile robots.
However, no such work addresses the Active Object Recognition of movable objects.
\\
In our approach, we present a solution to reliably recognize objects manipulated by a robot arm by learning poses in which an agent moves the robot arm's end effector to increase classification performance.
We use the Soft Actor-Critic, a method by Haarnoja et al. (2018) \cite{HAAR18sacaa}. 
It is an actor-critic DRL method with Soft Q-learning \cite{HAAR17entropy}, which is based on an entropy regularizer \cite{ZIE18mef} using the maximum entropy framework \cite{JAYNES57entropy1}.
Mahmood et al. (2018) \cite{MAHM18benchmarking} have made a benchmark of various robotic tasks, including multiple reinforcement learning algorithms, including the Soft Q-learning, which was rated the fastest learner.
In addition, they applied the double Q-learning trick and added an autoregulated entropy parameter \cite{HAAR18sacaa}. 
\begin{figure}
	\centering
	\includegraphics[width=0.9\columnwidth]{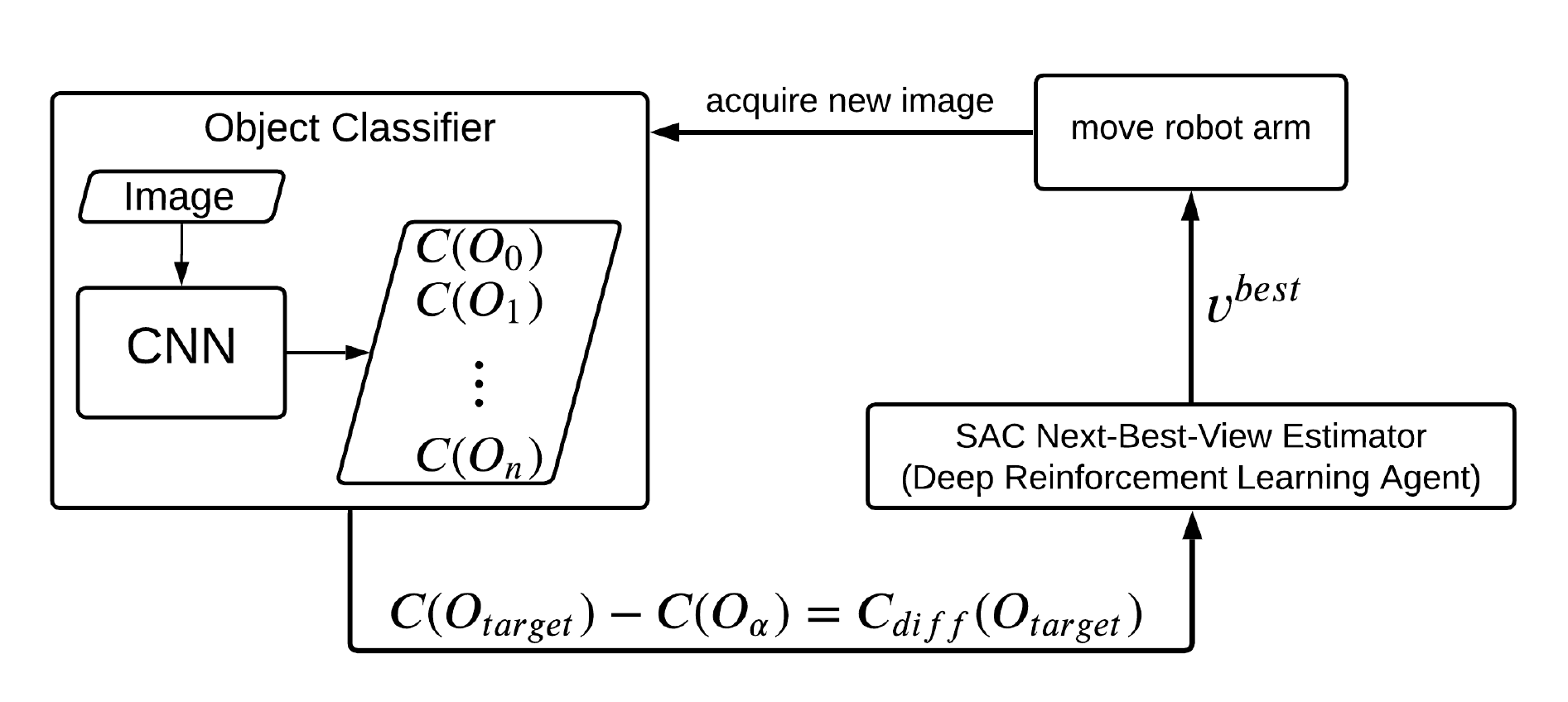}
	\caption{Interaction loop between object classifier and next-best-view estimator}
	\label{fig:A:loop}
\end{figure}

\section{Approach}
\label{sec:approach}
\subsection{Setup}\label{sec:A:setup}
\begin{figure}[H]
	\centering
	\subfloat[Simulation Environment]{{\includegraphics[width=0.4\columnwidth]{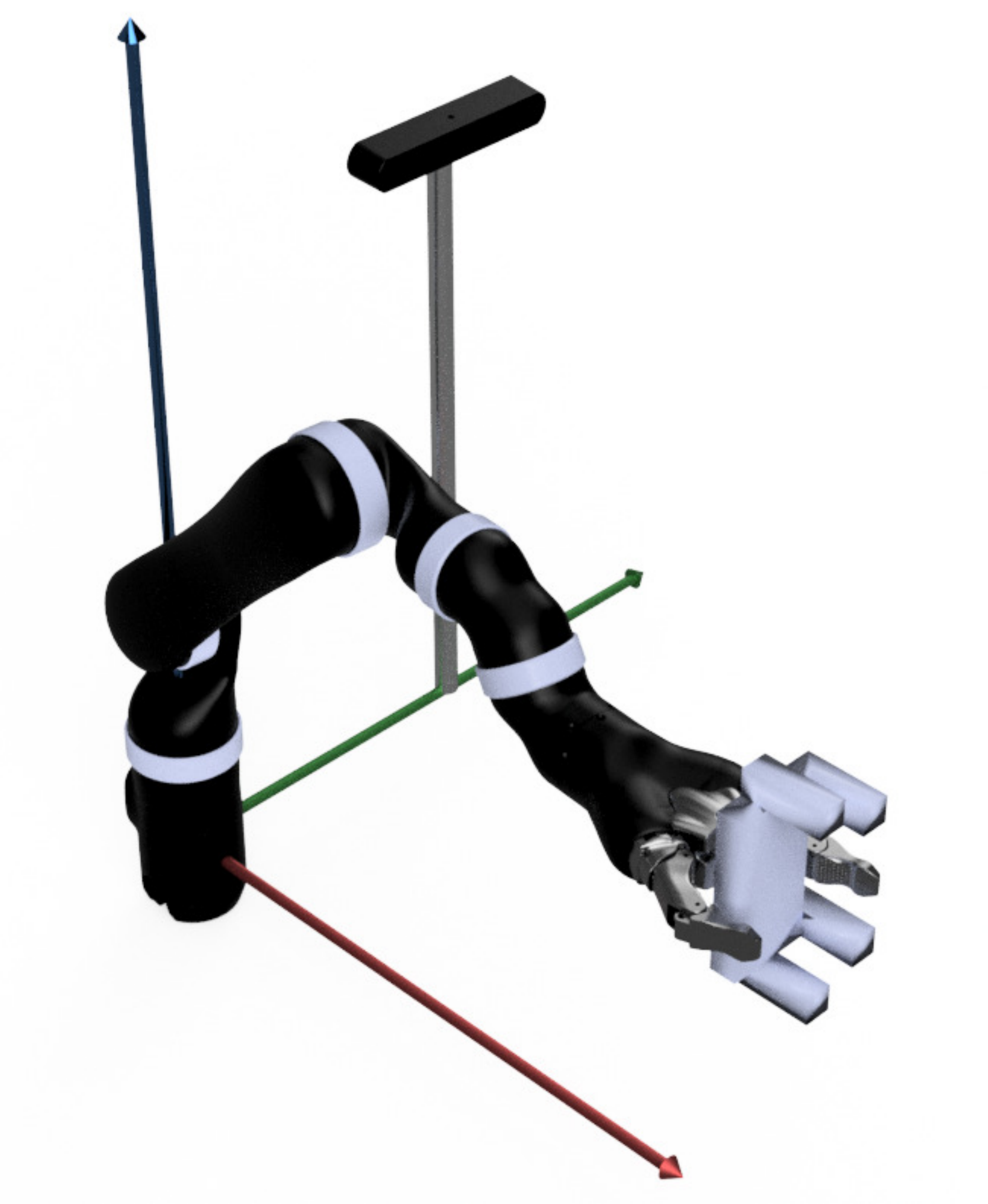} }\label{fig:OC:sim}}%
	\label{fig:OC:dd}
	\subfloat[TEOS building blocks]{{\includegraphics[width=0.5\columnwidth]{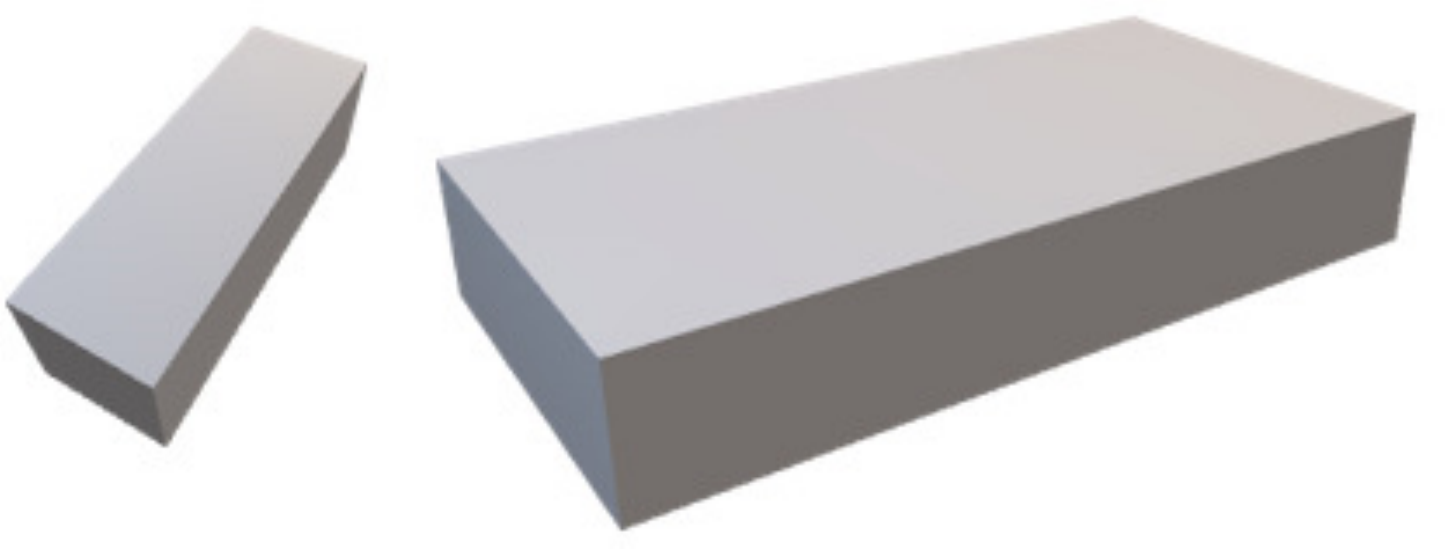}}\label{fig:OC:robot_data_set}}%
	\label{fig:OC:ee}
	\caption{Simulation environment setup, showing the robot coordinate system with the horizontally aligned stationary camera and an object in the end-effector. The  objects of TEOS are built on two building blocks; cuboid (left) and base (right) \cite{solbach2021blocks}.}
	\label{fig:A:setup}
\end{figure}

In this work, we present an approach to classify objects moved dynamically by a robot arm, while a camera is mounted statically.
This is the inverse of the problem that Wilkes and Tsotsos solved in \cite{WIL92cvpr}.
We propose an iterative process between object classification and next-best-view estimation with deep reinforcement learning, illustrated in Fig. \ref{fig:A:loop}.
The setup is based on a real-world setup consisting of a Kinova Gen2 7 DoF 1 robot arm for object manipulation and a static Stereolabs ZED mini 2 stereo camera for object classification. 
The components are simulated in PyBullet \cite{pybullet} with the exact specifications, including models and technical specifications.
We use a perfect lighting environment by disabling shadows so that we don't have to deal with self-shadowing, where the components of the object can cast shadows on themselves depending on the position of the light source.
The object classifier is implemented in PyTorch \cite{pytorch}, while the deep reinforcement learning algorithm is implemented in OpenGym \cite{gym} and Stable-Baselines \cite{stable-baselines}.
An abstraction of the simulation environment is shown in Fig. \ref{fig:A:setup}.
In the ideal process of object recognition including our setup, an optimal grasping point of the lying object would first be determined.
The optimal grasping point is intended to maximize the chance of identifying the object.
The grasping point should minimize the occlusion between the object and the arm, as well as the self-occlusion of the object.
After the following grasp, the pose with the highest information gain is determined, into which the object is moved next, until the  final pose is reached from which the most information can be obtained.
If the result is not sufficient or there are reachability issues of the grasping point, further grasping actions follow until the goal is reached.
\\
In this work, we ignore the grasping process and explore the performance and limitations of the subsequent object manipulation.
The moved object is further called \textit{target object} and remains in the same pose relative to the robot arm's end-effector during all experiments and evaluations, if not stated otherwise.
This positioning of the objects of TEOS will be explained in Section \ref{sec:A:DS}.
\\
The robot arm is placed at the origin $(0, 0, 0)$, while the camera has an horizontal viewing angle and is translated to $(0.003, 0.23, 0.53)$.
This position adjusts the camera's viewing frustum to the robot arm's workspace, where the robot arm can best move the objects to investigate them.
In contrast to the inspection of objects in open space, the viewpoint on our objects depends on the pose of our end-effector in the world space and the relative pose of the object to the end-effector.
The use of a robot arm reduces the object's distinctiveness, which is caused by two main reasons:

\paragraph{Occlusion}\label{sec:A:occlusion}
The end-effector, which holds the objects, reduces the visible area on the object.
This makes the classifier more prone to errors, especially if the objects used already have a lot of self-occlusion.

\paragraph{Reachability } 
The limited freedom of movement of the robot arm may not allow it to reach the target position.
This problem strongly depends on the configuration and positioning of the camera and robot arm.
Also, it is less likely to occur with higher degrees of freedom (DoF) robot arms.
\par

\subsection{Dataset}\label{sec:A:DS}
\begin{figure}[H]
		\vspace{0.08in}
	\centering
	\includegraphics[width=0.5\columnwidth]{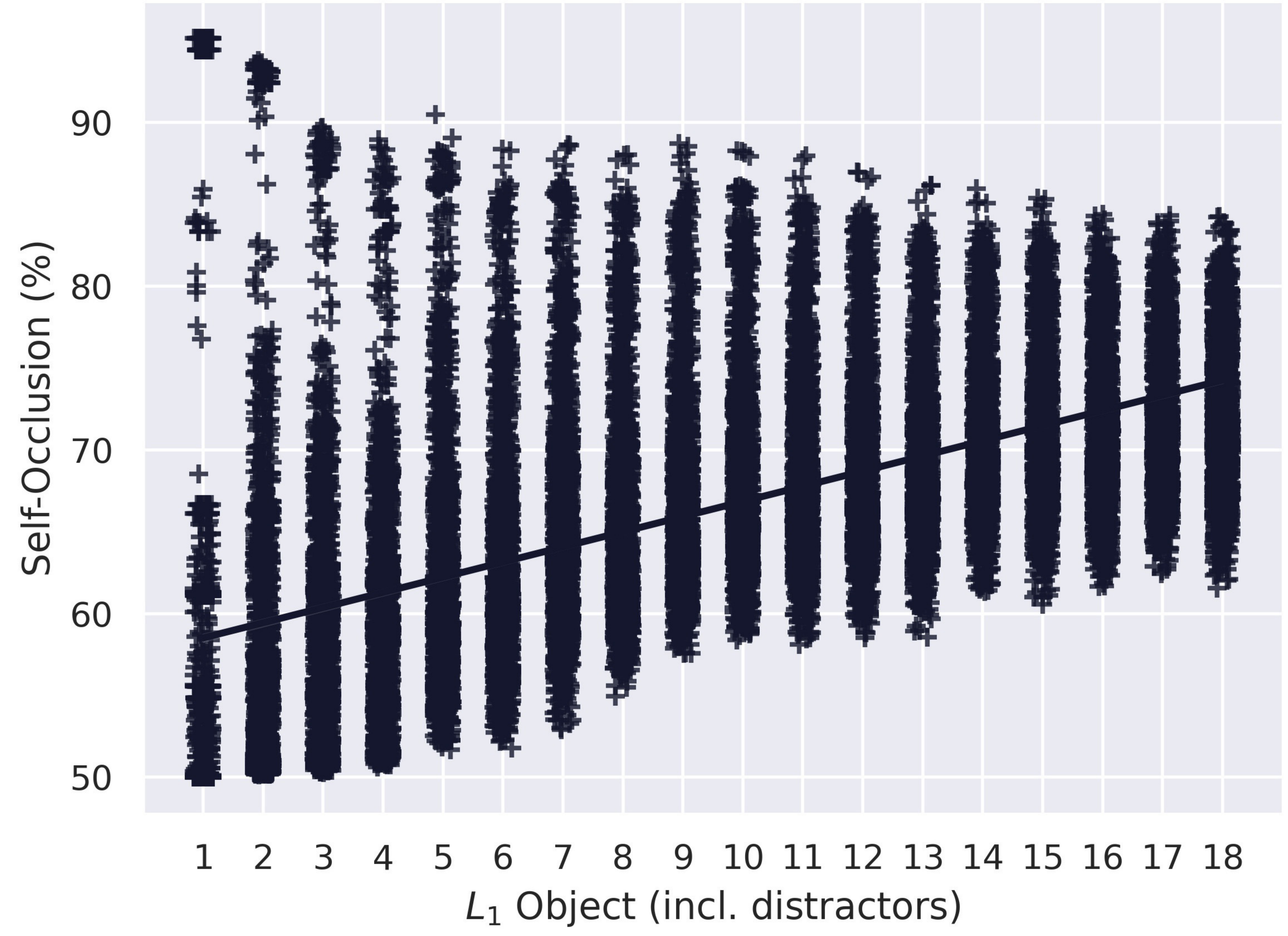}
	\caption{Illustration of the amount of self-occlusion per object of the TEOS $L_1$ dataset \cite{solbach2021blocks}. Each point shows the self-occlusion of the respective object from a specific viewpoint. The viewpoints are evenly distributed on a sphere around an object, resulting in 768 unique views. The straight line illustrates the increase in average self-occlusion as the complexity increases.}
	\label{fig:A:teos_fig12}
\end{figure}

\begin{figure}[H]
	\centering
	\includegraphics[width=0.5\columnwidth]{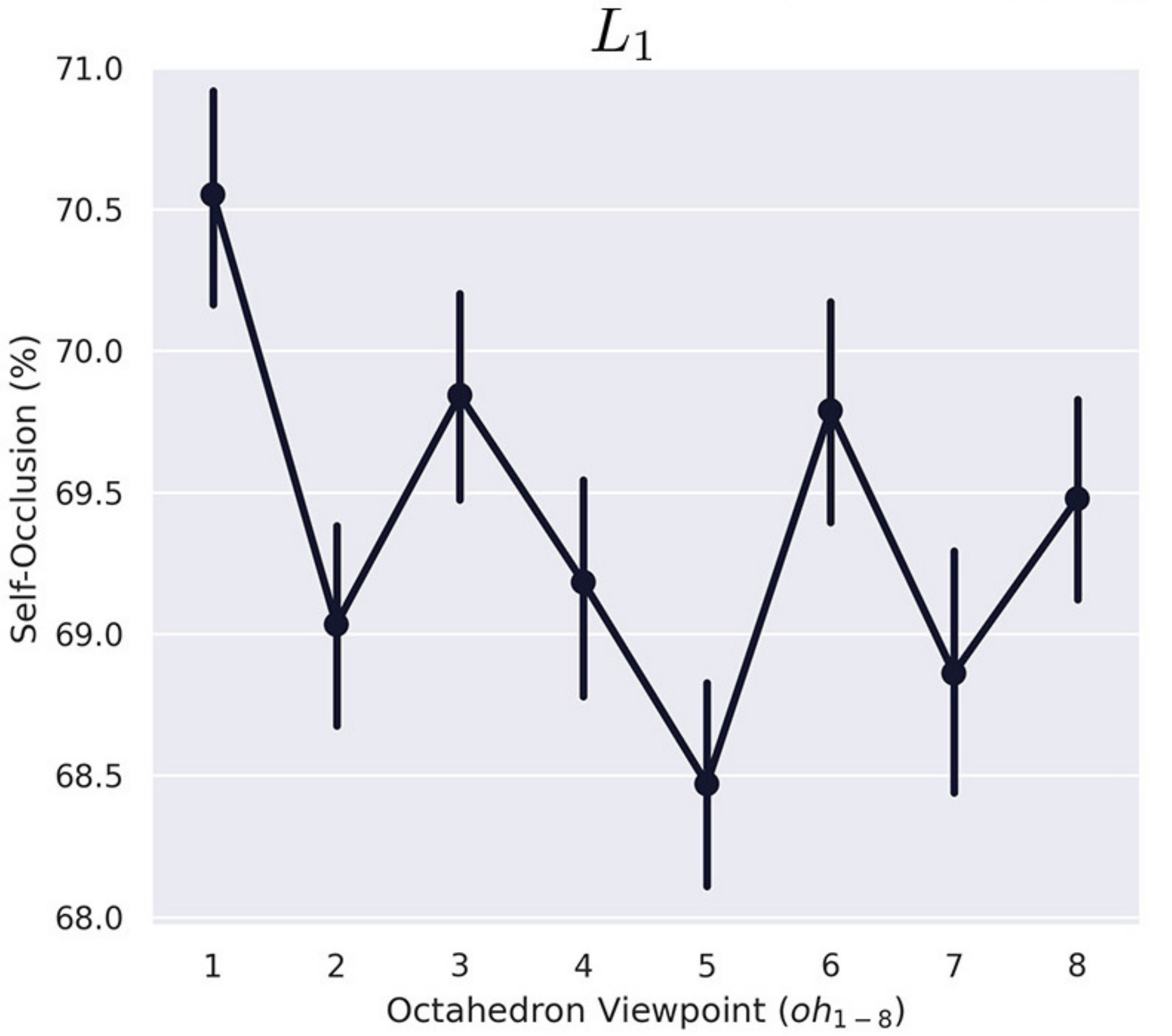}
	\caption{Illustration of the distributional relation between viewpoint
		mapping and self-occlusion for the TEOS $L_1$ dataset. The octahedron viewpoint ($oh_{1-8}$) stands for all points projected onto one of 8 sectors that comprise the complete viewing sphere. The dots show the average for all viewpoints located in the respective octahedron. The vertical line shows the 95\% confidence interval. 
		For further information please see \cite{solbach2021blocks}.}
	\label{fig:A:teos_fig17}
\end{figure}

The TEOS dataset \cite{solbach2021blocks} includes 48 objects splitted into two sets; $L_1$ and $L_2$. $L_1$ consists of 36 objects in 18 complexity levels, $L_2$ of 12 objects in three complexity levels.
A complexity level depends on the structure of the object.
We use $L_1$, where each object is made of a base and a fixed number of cuboids $j \textnormal{ with } 0 <= j < 17$, illustrated in \ref{fig:A:setup}.
The objects are noted as $O_k$  with $k = j + 1$ .
In addition, for each object there is an identical distractor object, which differs only by a differently oriented cuboid.
In this paper, we exclude the distractors, so that we end up with 18 unique objects representing 18 complexities, computed as $compl = k = j + 1$. 
We further name this reduced dataset $L_1^{objects}$.
Image samples are shown in Fig. \ref{fig:OC:object_data_set}.
\\
We specifically chose TEOS due to its known complexity and self-occlusion levels. 
The self-occlusions describes the occlusion caused by the target object to itself.
Fig. \ref{fig:A:teos_fig12} shows the increasing amount of self-occlusion with respect to object complexity, while the self-occlusion distribution per class decreases. 
However, as pointed out by the authors of TEOS, the viewpoint plays a direct role in the observed self-occlusion. 
As shown in Fig. \ref{fig:A:teos_fig17}, certain viewpoints mitigate the amount of self-occlusion.
\par

\subsection{Object Classifier}\label{sec:A:OC}
\begin{figure}[H]
	\centering
	\subfloat[Unit component]{{\includegraphics[width=0.8\columnwidth]{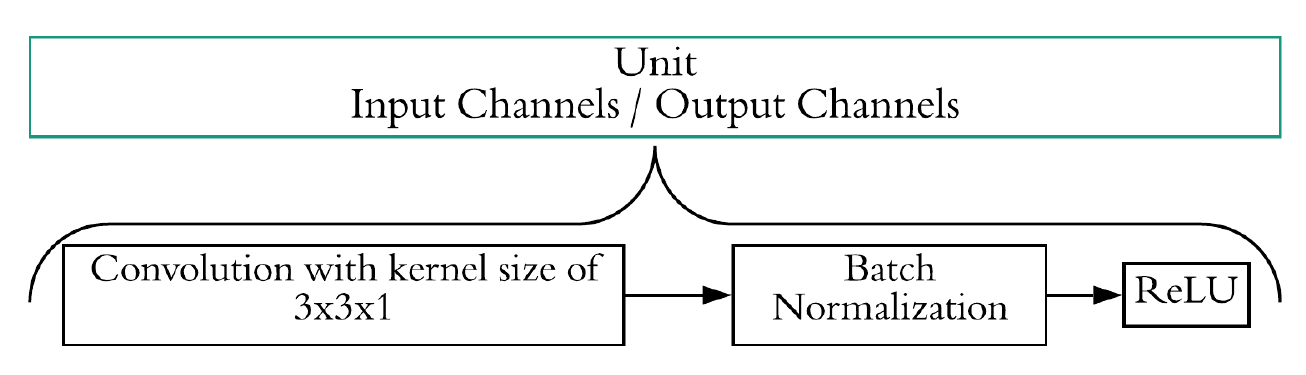} }\label{fig:A:OC:basicnet_unit}}%
	\label{fig:OC:d}
	\subfloat[Neural network architecture]{{\includegraphics[width=1.0\columnwidth]{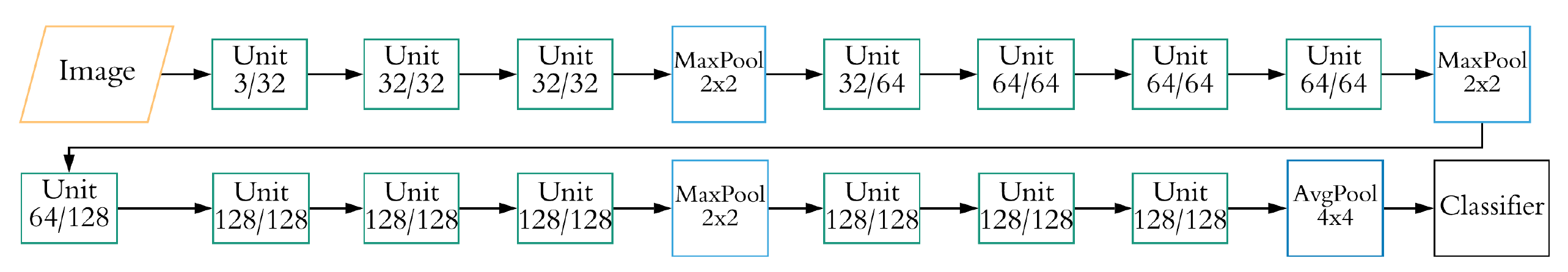}}\label{fig:A:OC:basicnet_architecture}}%
	\label{fig:OC:e}
	\caption{Object classifier architecture.}
	\label{fig:A:OC:basicnet}
\end{figure}

To classify our dataset, we use a simple deep convolutional neural network (CNN) architecture, since for TEOS, the most simple CNN performed the best \cite{solbach2021blocks}.
The CNN is based on the approach of J. Olafenwa \cite{basicnet}.
We define a processing unit, illustrated in Fig. \ref{fig:A:OC:basicnet_unit}, as a convolutional layer with given input and output channels followed by a batch normalization layer and ReLU activation layer.
The convolutional layer has a kernel size of $3$, the striding and padding a dimension of $1$.

Therefore, the whole neural network is made up of 14 convolution layers, 14 batch normalization layers, 14 ReLU layers, 4 pooling layers consisting of 3 max-pooling layers and 1 average pooling, and 1 fully connected layer, totaling 62 layers, shown in Fig. \ref{fig:A:OC:basicnet_architecture}.
The last linear layer ensures the classification of our $n$ classes based on 128 input features.

\subsection{Next-Best-View Estimator}\label{sec:A:NBV}
Given a dataset with objects that cannot be distinguished from particular viewpoints, we want to find a sequence $seq_v$ of $n$ viewpoints $v_n$ that provide additional and needed information to improve classification.
A sequence consists of single viewpoints, called next-best-views.
The sequence ends when the classification confidence reaches a defined threshold.
In the ideal case, the first chosen next-best-view leads to the pose with the highest classification confidence directly.
\\
Instead of using confidence as a classification metric, we specify an extended metric, called \textit{confidence difference}.
First, we define the object $O_\alpha$ as the object with the highest classification confidence besides the target object $O_{target}$, so that $O_{target} \neq O_\alpha$.
Let $C(O_{target})$ be the classification confidence of the target object $O_{target}$ and $C(O_\alpha)$ the classification confidence of object $O_\alpha$.
Then the classification confidence difference is defined as 
\begin{eqnarray}
C_{diff}(O_{target}) = C(O_{target}) - C(O_\alpha).
\end{eqnarray}
This results in a negative value $C_{diff}(O_{target}) < 0$ if the target object is not the best recognized object.
This value, called \textit{confidence difference} in the following, is the most important value.
It expresses more than the usual confidence of our target object.
An confidence of 51\% does not mean that we can clearly distinguish the target object if another object has an confidence of 49\%, which results in an confidence difference of 2\%.
However, if the prediction of all other objects is more evenly distributed, we get an higher confidence difference and thus, a more explicit prediction.
\\
We now define the goal of finding the next-best-view $v_{best}$ for object $O_k$ as
\begin{equation}
	v^{best} = \arg\max\limits_{v} C_{diff}^v(O_k),
\end{equation}
where $v_{O_k}^{best}$ is a 6 DOF pose of the robots end-effector holding object $O_k$.
We define the sequence ${seq}$ of $n$ next-best-views as
\begin{equation}
{seq}_v = v_0^{best}, v_1^{best}, \dots, v_n^{best}, n >= 0.
\end{equation}
\begin{figure*}[h] 
	\vspace{0.06in}
	\centering$
	\begin{array}{ccc}
	\includegraphics[width=.31\linewidth]{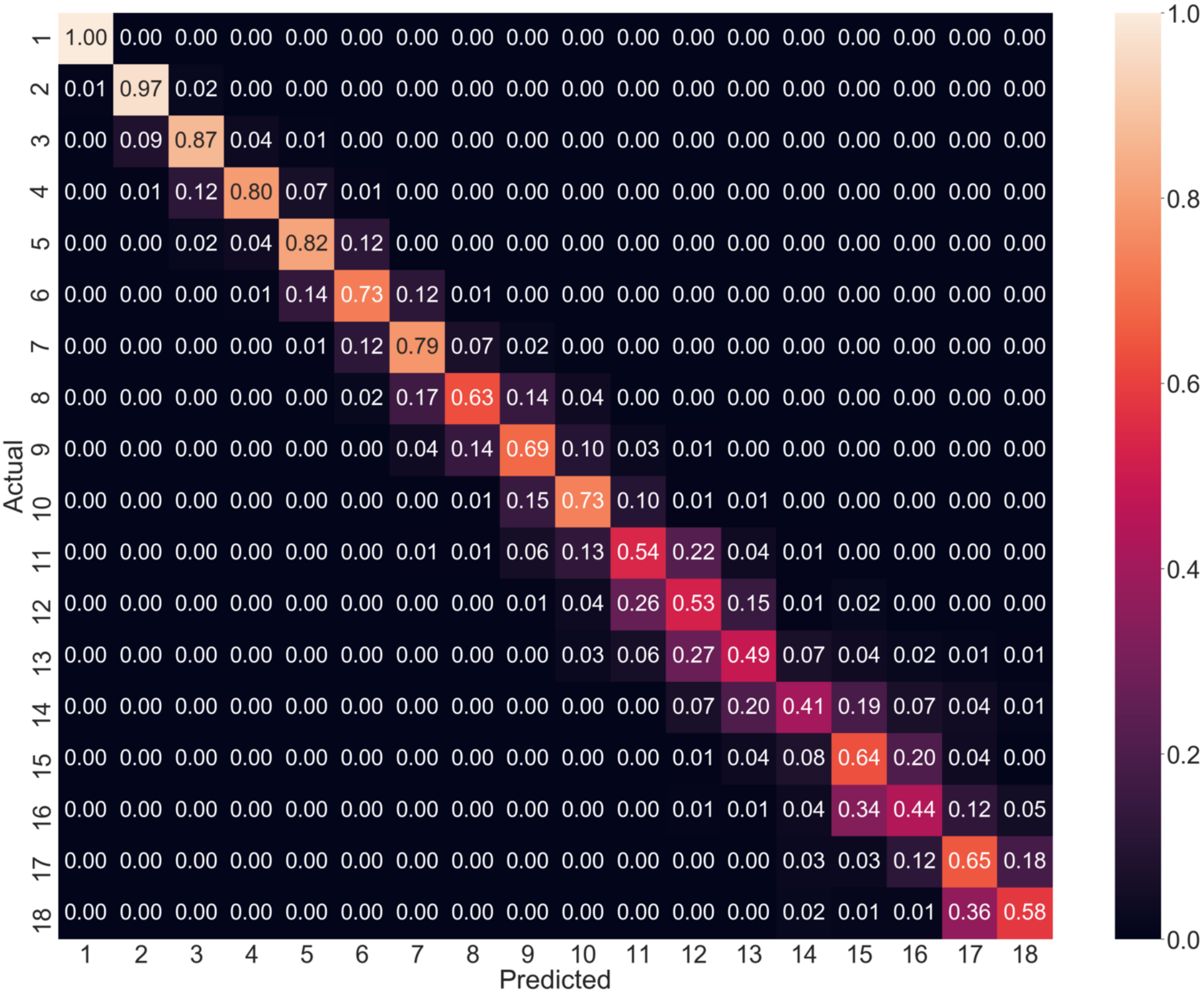} &
	\includegraphics[width=.31\linewidth]{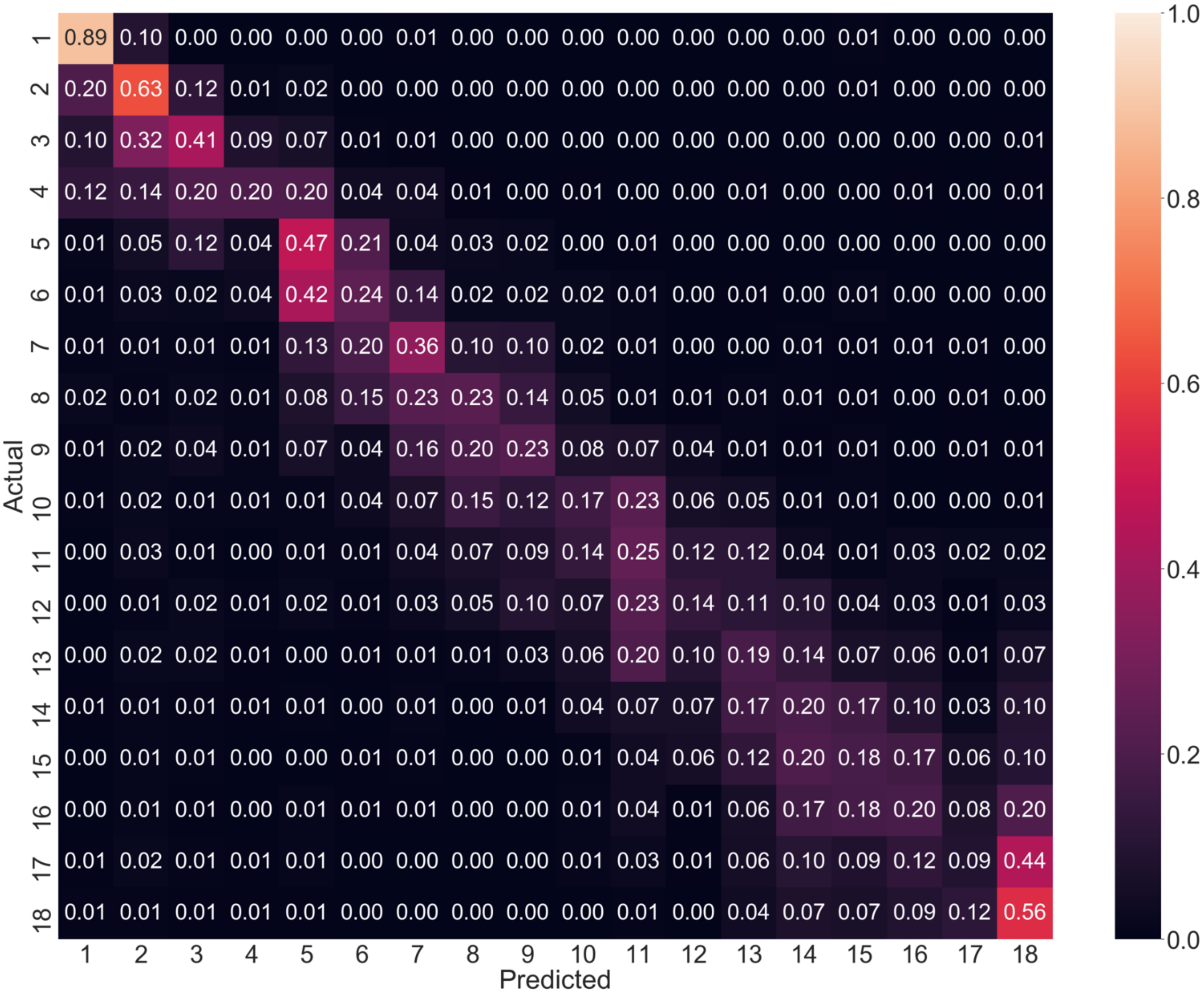} &
	\includegraphics[width=.31\linewidth]{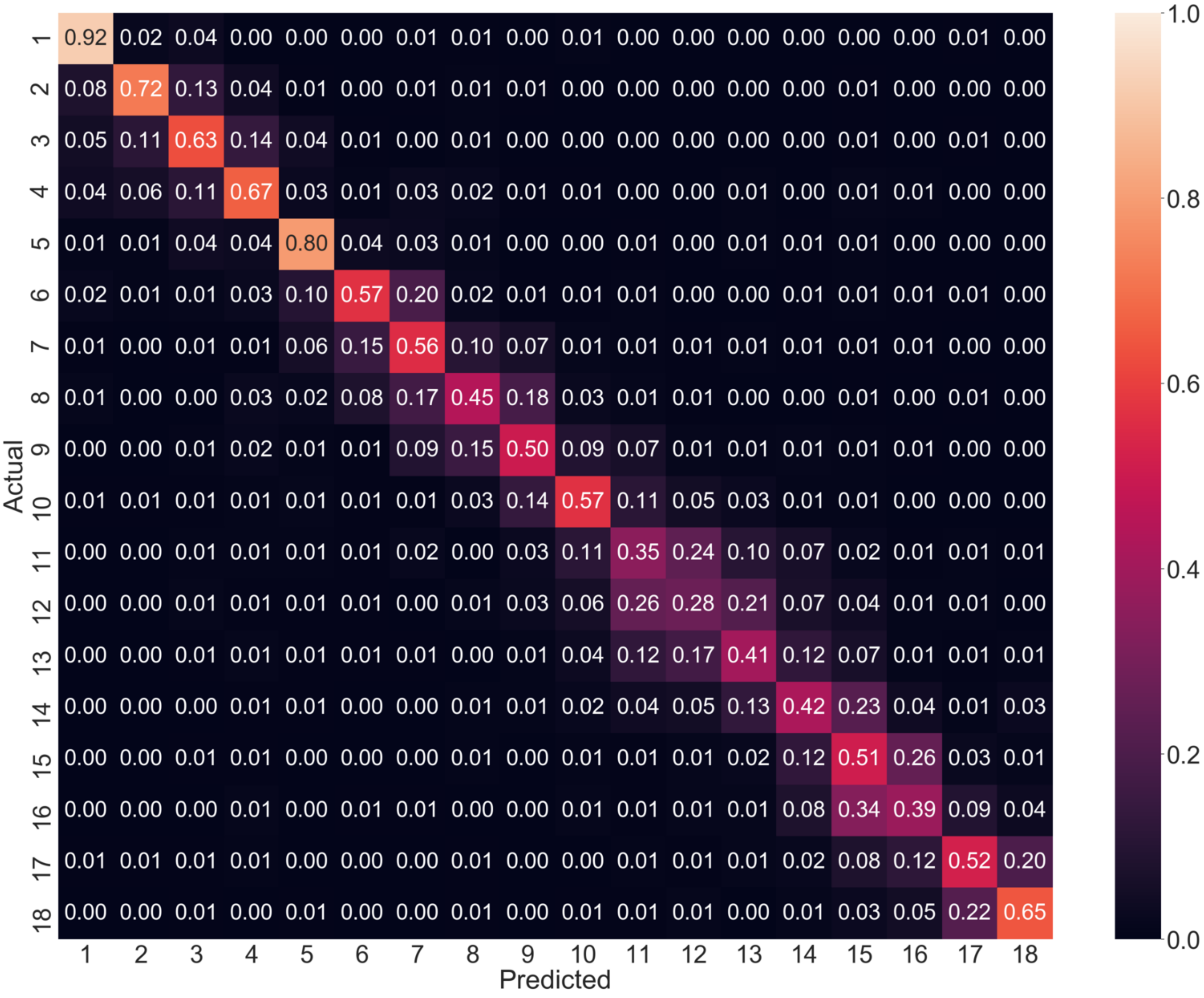}
	\\
	\textrm{$M(S_{objects})$ on $S_{objects}$} & \textrm{$M(S_{robot})$ on $S_{robot}$}  & \textrm{$M_{pre}(S_{robot})$ on $S_{robot}$} 
	\end{array}$   \caption{Confusion matrices of the classification models on specific image sets.}
	\label{fig:EXP:cm_with_arm}%
\end{figure*}
We use the \textit{Soft Actor-Critic} (SAC) \cite{HAAR18sacaa}, a deep reinforcement learning algorithm, to train an agent that finds the sequence of next-best-views by learning the poses that deliver the highest classification confidence difference, visualized in Fig. \ref{fig:A:loop}.
The algorithm is, based on its benchmarks, one of the most efficient model-free algorithms and mainly designed for real-world robotic systems to solve tasks ranging from manipulation to locomotion.
The used SAC hyperparamters are listed in Table \ref{tab:SAC:params}.
\\
We move the robot arm with its object by inputting a pose with an orientation given as a quaternion. 
Therefore, we are moving in Cartesian space and define a continuous box with predefined boundaries for the action space.
The boundaries define a 3D region in front of the camera and the observation space consists of two components.
The first is the pose in which the end-effector is positioned, the second component is the confidence difference.
The reward is also computed by the confidence difference defined by $r = C_{diff}(O_{target})$.
Ideally, the pose is identical to the one chosen by the agent as the action but may differ if the robot arm cannot fully reach the pose.

\section{Experiments}
\label{sec:experiments}
\subsection{Object Classifier}\label{sec:EXP:OC}
\begin{figure}
	\centering
	\subfloat[Object-only Set $S_{objects}$]{{\includegraphics[width=0.45\columnwidth]{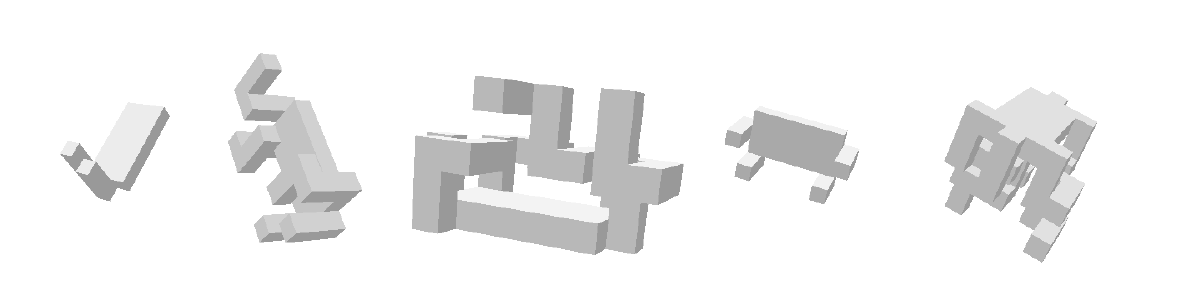} }\label{fig:OC:object_data_set}}%
	\label{fig:OC:d}
	\subfloat[Robot Set $S_{robot}$]{{\includegraphics[width=0.45\columnwidth]{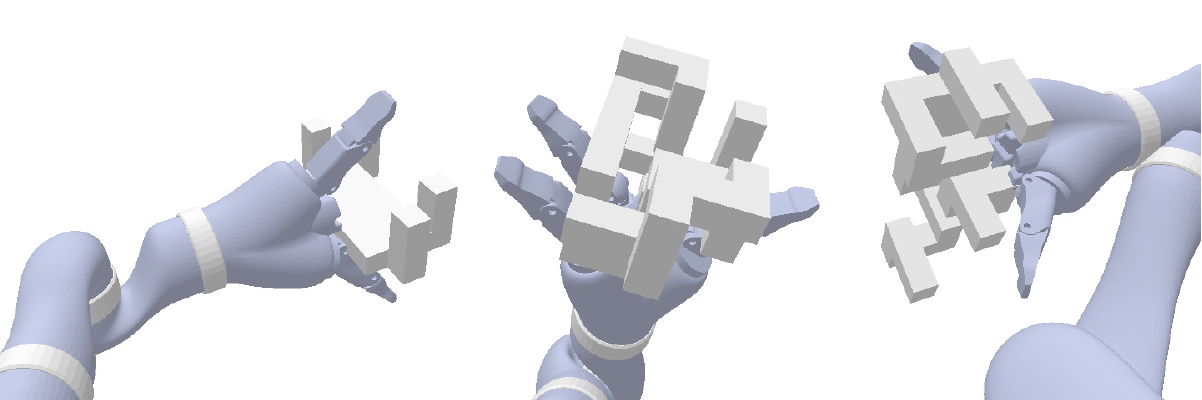}}\label{fig:OC:robot_data_set}}%
	\label{fig:OC:e}
	\caption{Sample images}
	\label{fig:OC:data_set}
\end{figure}

To begin, we generate two image sets, $S_{objects}$ and $S_{robot}$, to evaluate the performance of $L_1^{objects}$ explained in Section \ref{sec:A:DS} applied to the robot arm setup.
$S_{robot}$ consists of images where the robot arm holds the objects, while 
$S_{objects}$ only shows the objects alone, excluding the arm.
For each image sets, we generate 2000 random poses respectively and create an image for each object-pose combination.
We split each image set in a train, validation, and test set by following the $80\%/10\%/10\%$ principle, though every created image set consists of the same poses.
Sample images are shown in Fig. \ref{fig:OC:data_set}.
\\
We use the neural network explained in Section \ref{sec:A:OC} to generate three different classification models based on the reduced $L_1^{objects}$ TEOS dataset explained in Section \ref{sec:A:DS}.
The first classification model $M(S_{objects})$ is trained on $S_{objects}$, the second one $M(S_{robot})$ on $S_{robot}$.
$M(S_{objects})$ can reliably predict most objects of the $S_{objects}$ image set, but the prediction is unsuccessful for $S_{robot}$. 
$M(S_{robot})$ fails to produce significant values for both image sets.
To achieve sufficient performance for the robot image set, we train an additional classification model $M_{pre}(S_{robot})$ using $M(S_{objects})$ as a pre-trained model.
This model can reasonably recognize the robot image set.

\begin{figure}
	\centering
	\subfloat[$M(S_{objects})$ on $S_{objects}$]{{\includegraphics[width=0.65\columnwidth]{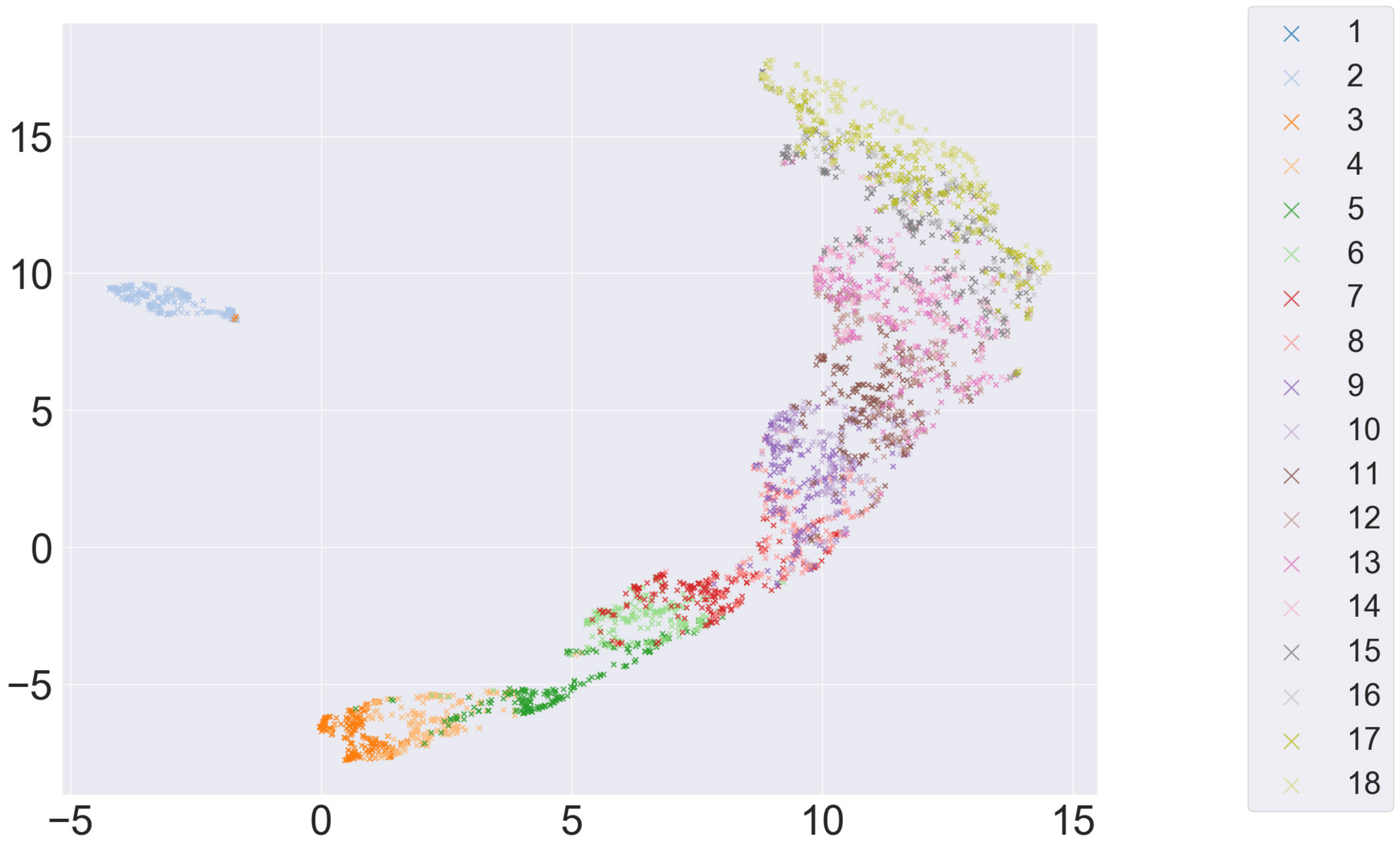} }}%
	\qquad
	\subfloat[$M_{pre}(S_{robot})$ on $S_{robot}$ ]{{\includegraphics[width=0.65\columnwidth]{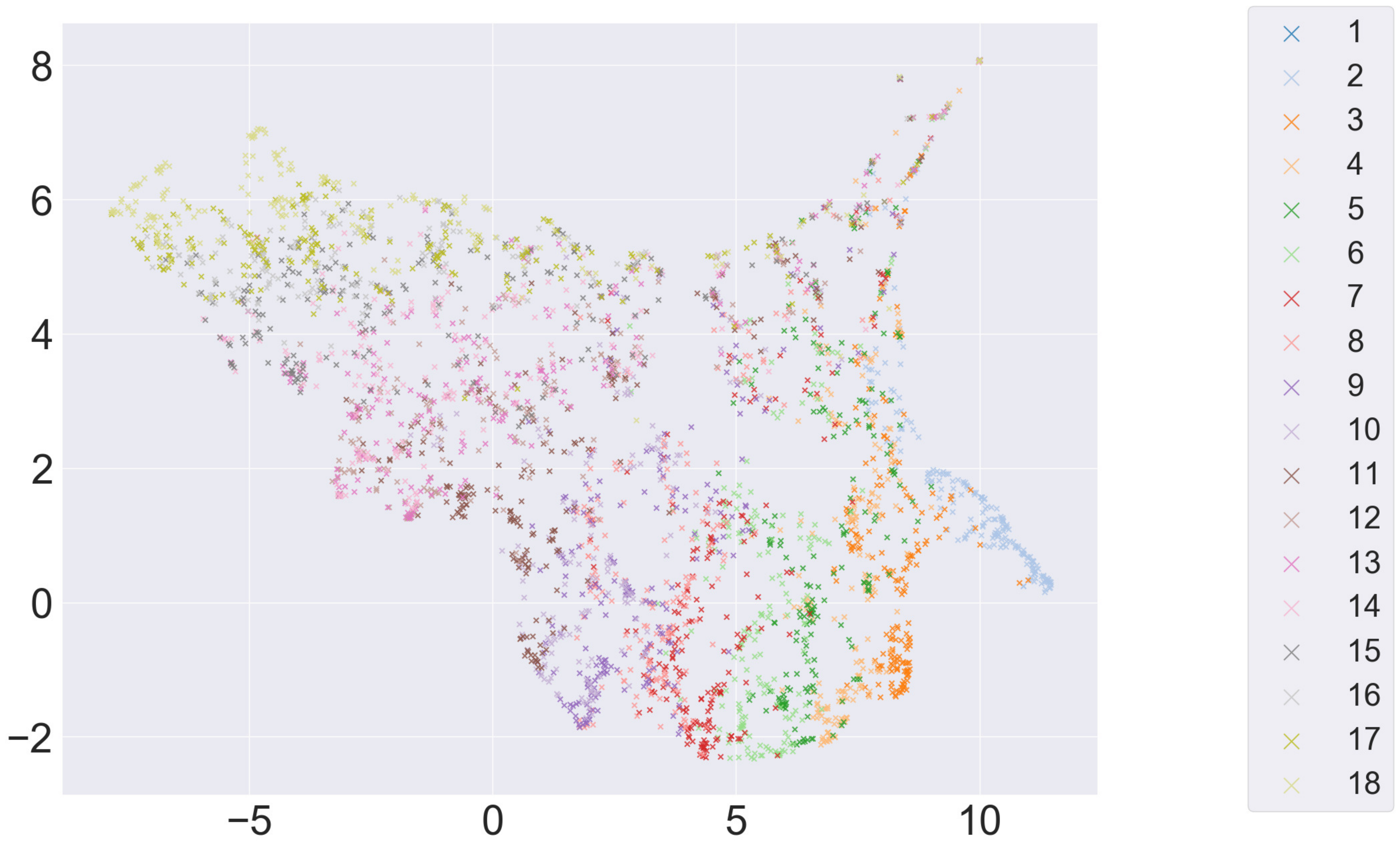}}}
	\caption{Dimension reduction embedding (UMAP).}
	\label{fig:EXP:cm_with_arm_umap}
\end{figure}

The performances of all three models are stated in Table \ref{tab:OC:prediction} and visually presented by confusion matrices in Fig. \ref{fig:EXP:cm_with_arm}.
The figure shows that correct prediction becomes easier the lower the complexity.
\\
We can recognize the complexity of classifying the reduced TEOS dataset in context with the robot arm holding the objects by using a dimension reduction technique, the \textit{Uniform Manifold Approximation and Projection} (UMAP) \cite{mcinnes2018umap-software}, visualized as embedding in Fig. \ref{fig:EXP:cm_with_arm_umap}.
The embedding initially shows that the objects can not easily be separated due to their similar views and lack of additional view information.
We can see strongly structured clusters for model $M(S_{objects})$, where the objects are arranged in ascending order of complexity and mostly have 1-2 adjacent classes, with the exception of object 1, which is encapsulated and consists of no cuboids other than the base.
But we get a broader distribution for $M_{pre}(S_{robot})$, where the transitions between the objects are smoother than in the former, resulting in more overlapping and multiple adjacent object classes.
\\
Additionally, since there is an average accuracy of 55.11\% for $M_{pre}(S_{robot})$, as shown in Table \ref{tab:OC:prediction}, it is not sufficient to always classify objects reliably, we need better viewpoints on the respective objects.
In Fig. \ref{fig:EXP:cm_with_arm}, we can especially see that for $M_{pre}(S_{robot})$ some objects are below 50\%.
Thus, we need sequences of next-best-views in order to ensure a successful object classification of these objects.
\begin{table}
	\renewcommand{\arraystretch}{1.15}
	\centering
	\caption{Average prediction accuracy of the used classification models on different image datasets.}
	\label{tab:OC:prediction}
	\begin{tabular}{c|c|c}
		\hline
		\bfseries Model & \bfseries Object-only dataset & \bfseries Robot dataset\\\hline
		$M(S_{objects})$ & 68.39 & 06.22\\
		$M(S_{robot})$ & 27.94 & 31.33\\
		$M_{pre}(S_{robot})$ & 21.28 & 55.11\\
		\hline
	\end{tabular}
\end{table}

\subsection{SAC Next-Best-View Estimator}
\begin{figure}
	\centering
	\subfloat[Trained agents.]{{\includegraphics[width=0.55\columnwidth]{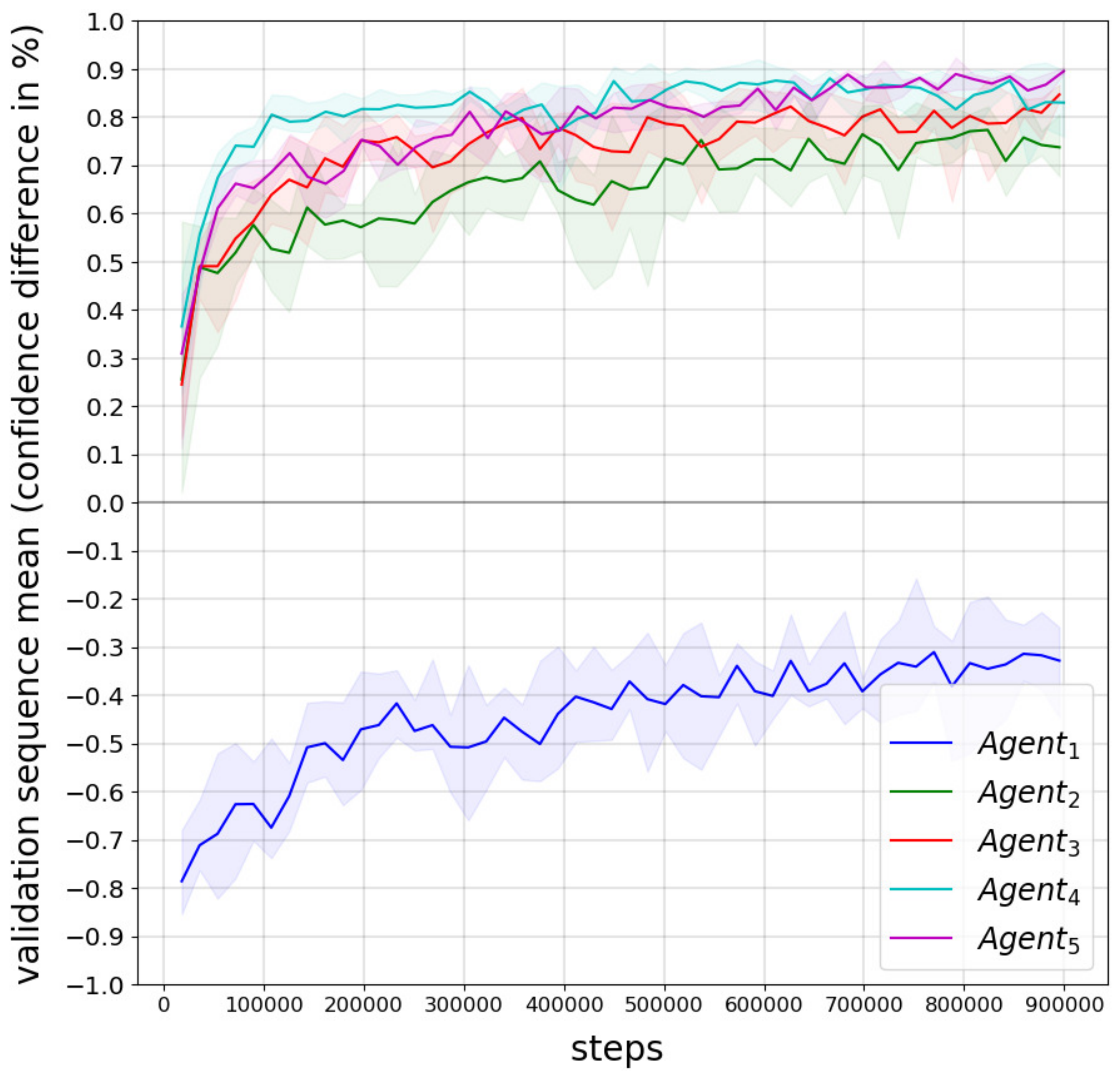} }\label{fig:EXP:5_val_mean}}%
	\qquad
	\subfloat[$Agent_5$ for certain objects.]{{\includegraphics[width=0.55\columnwidth]{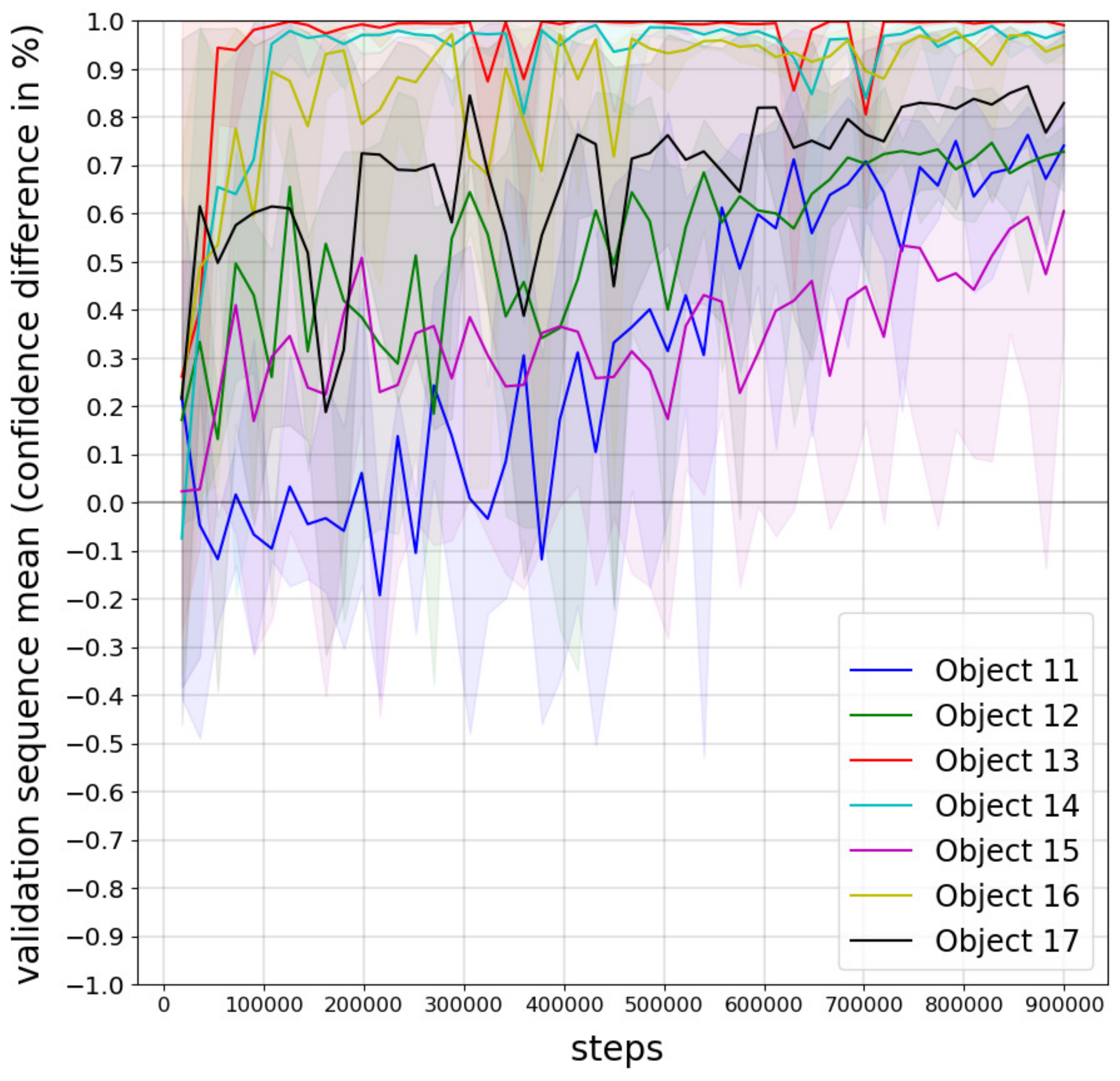}}\label{fig:EXP:6_val_mean}}
	\caption{Validation of the sequence means.}
	\label{fig:EXP:val_mean}
\end{figure}

We define the default environment setup including the robot arm as robot mode $m_{robot}$ and an additional setup excluding the robot arm as object-only mode $m_{objects}$.
In $m_{objects}$, the objects are not connected to the end-effector and do not depend on the robot arm's movement.
We run the agent's training for a total step size of 900,000 and execute a validation each 18,000 steps.
The validation consists of 5 episodes per object, in which we validate the confidence difference mean of a sequence of 10 steps.
Fig. \ref{fig:EXP:5_val_mean} illustrates the validation results for all different specifications we used in this work.
SAC learning begins after each object has been trained for a total of 1,000 steps, resulting in initial validation based only on random poses to gain additional information.

First, the two agents, $Agent_1$ and $Agent_2$, are trained in $m_{robot}$ by using classification model $M(S_{objects})$ and $M_{pre}(S_{robot})$ respectively, setting the step size per episode to 100.
The specifications of all agents is listed in \ref{tab:agents}.
$Agent_1$ can't achieve positive confidence difference values, what was expected based on the evaluation of the classifier on $m_{robot}$ with an average prediction accuracy of 6.22\% (see Table \ref{tab:OC:prediction}).
In contrast, $Agent_2$ achieves an average confidence difference of over 70\%.
\\
Since we have to deal with the occlusion of the robot arm relative to the objects, we also train a reference $Agent_3$ in mode $m_{objects}$ to explore the performance difference between the two different modes.
The new agent achieves a slightly increased average performance up to 10\%, thus, shows the added difficulty caused by the occlusion the system has to deal with.
Another agent $Agent_4$ is trained by reducing the steps per episode to 10, since we aim to achieve the best pose within the first actions.
This results in a noticeable improvement of the prediction up to 10\% compared to the previous $Agent_3$.
Especially the efficiency at the beginning of the training is strongly improved.
However, an even smaller step size did not empirically show any additional advantages, so we set the step size per episode to 10 for all subsequent experiments.
\\
$Agent_4$ is now being used as a pre-model for the new agent $Agent_5$ trained in mode $m_{robot}$ using $M_{pre}(S_{robot})$.
By using the pre-model we achieve further performance improvements.
$Agent_5$ achieves improvements of up to 20\% for the average validated confidence difference in mode $m_{robot}$ compared to $Agent_2$.
Fig. \ref{fig:EXP:5_val_mean} shows that $Agent_5$ can even outperform $Agent_4$, although the average prediction accuracy is about 13 \% lower (see Table \ref{tab:OC:prediction}).
\\
We now look at a coherent group of the validation of $Agent_5$.
We choose objects $O_{11}$ to $O_{17}$, which are difficult to detect by $M_{pre}(S_{robot})$.
These show values between 28\% and 52\% in the corresponding confusion matrix (Fig. \ref{fig:EXP:cm_with_arm}).
In Fig. \ref{fig:EXP:6_val_mean}, we recognize that $O_{11}$ starts in the negative range and only slowly begins to learn.
$O_{12}$ and $O_{13}$ already reach an classification confidence difference of almost 1 after 100,000 steps. 
The worst classified objects $O_{11}$ and $O_{12}$ end up with approximately 73\%.
The validation of some objects varies a lot at the beginning, but stabilizes towards the end.
Overall, all shown objects achieve a minimum of over 50\%, but no correlations can be found between the values from the confusion matrix and the validations.
\begin{table}
	\renewcommand{\arraystretch}{1.15}
	\centering
	\caption{Specifications of the trained agents}
	\begin{tabular}{c|c|c|c|c}
		\multirow{2}{*}{\bfseries Agent} & \multirow{2}{*}{\begin{tabular}[c]{@{}l@{}}\bfseries Classification\\ \bfseries Model\end{tabular}} & \multirow{2}{*}{\bfseries Mode} & \multirow{2}{*}{\begin{tabular}[c]{@{}l@{}}\bfseries Pre-trained\\ \bfseries Agent\end{tabular}} & \multirow{2}{*}{\begin{tabular}[c]{@{}l@{}}\bfseries steps per\\ \bfseries episode\end{tabular}} \\
		&                              &         &                                                                           &                                \\\hline
		1     & $M(S_{objects})$             & $m_{robot}$     &  -                 & 100 \\
		2     & $M_{pre}(S_{robot})$         & $m_{robot}$     &  -                 & 100\\
		3     & $M(S_{objects})$             & $m_{objects}$   &  -                 & 100 \\
		4     & $M(S_{objects})$             & $m_{objects}$   &  -                 & 10 \\
		5     & $M_{pre}(S_{robot})$         & $m_{robot}$     &  4                 & 10 \\\hline    
	\end{tabular}
	\label{tab:agents}
\end{table}
\subsection{Results}
\begin{figure}
	\centering
	\subfloat[Dimension reduction embedding (UMAP).]{{\includegraphics[width=0.8\columnwidth]{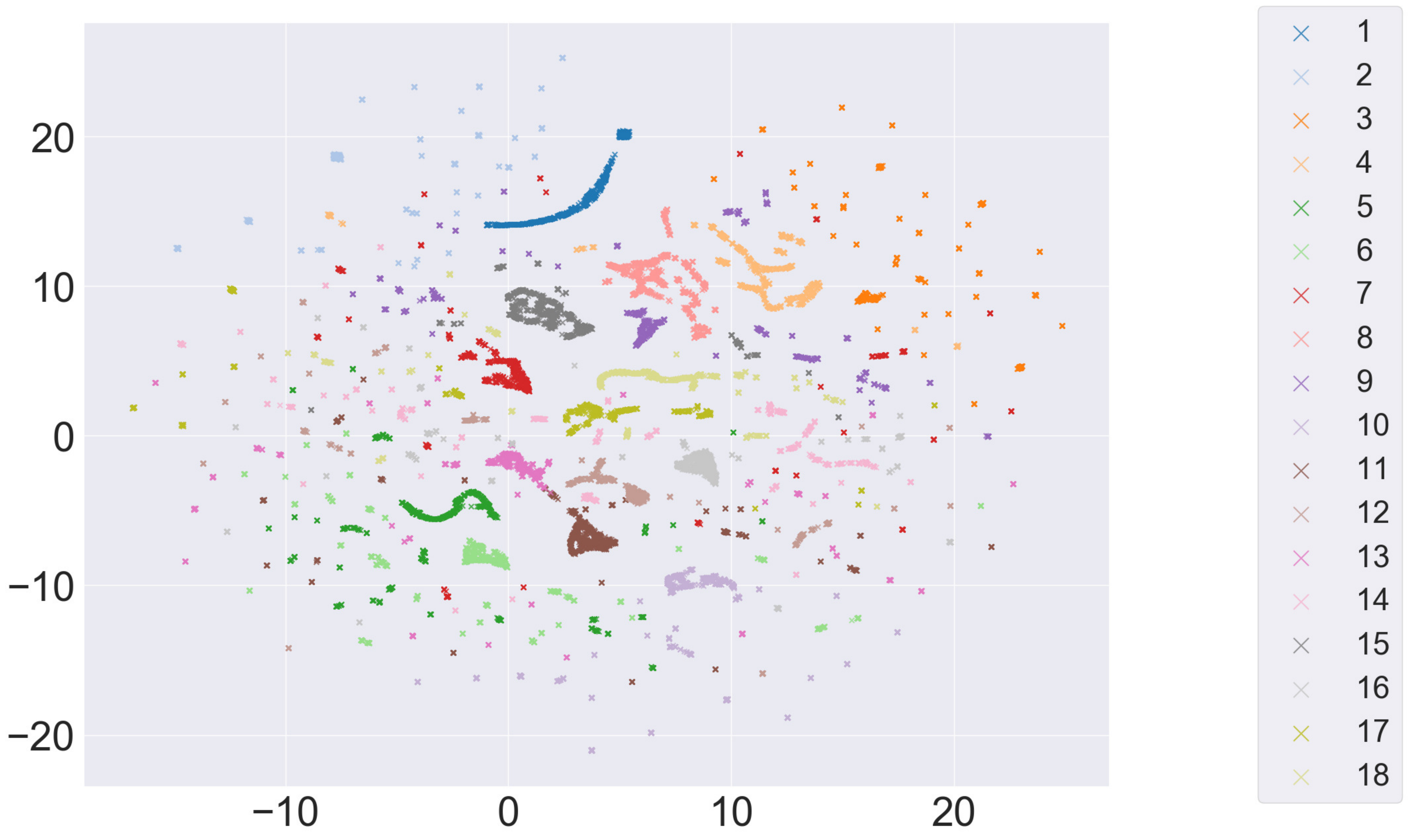} }\label{fig:EVAL:agent_umap}}%
	\qquad
	\subfloat[Confusion matrix with an accuracy of 96,33\%.]{{\includegraphics[width=0.7\columnwidth]{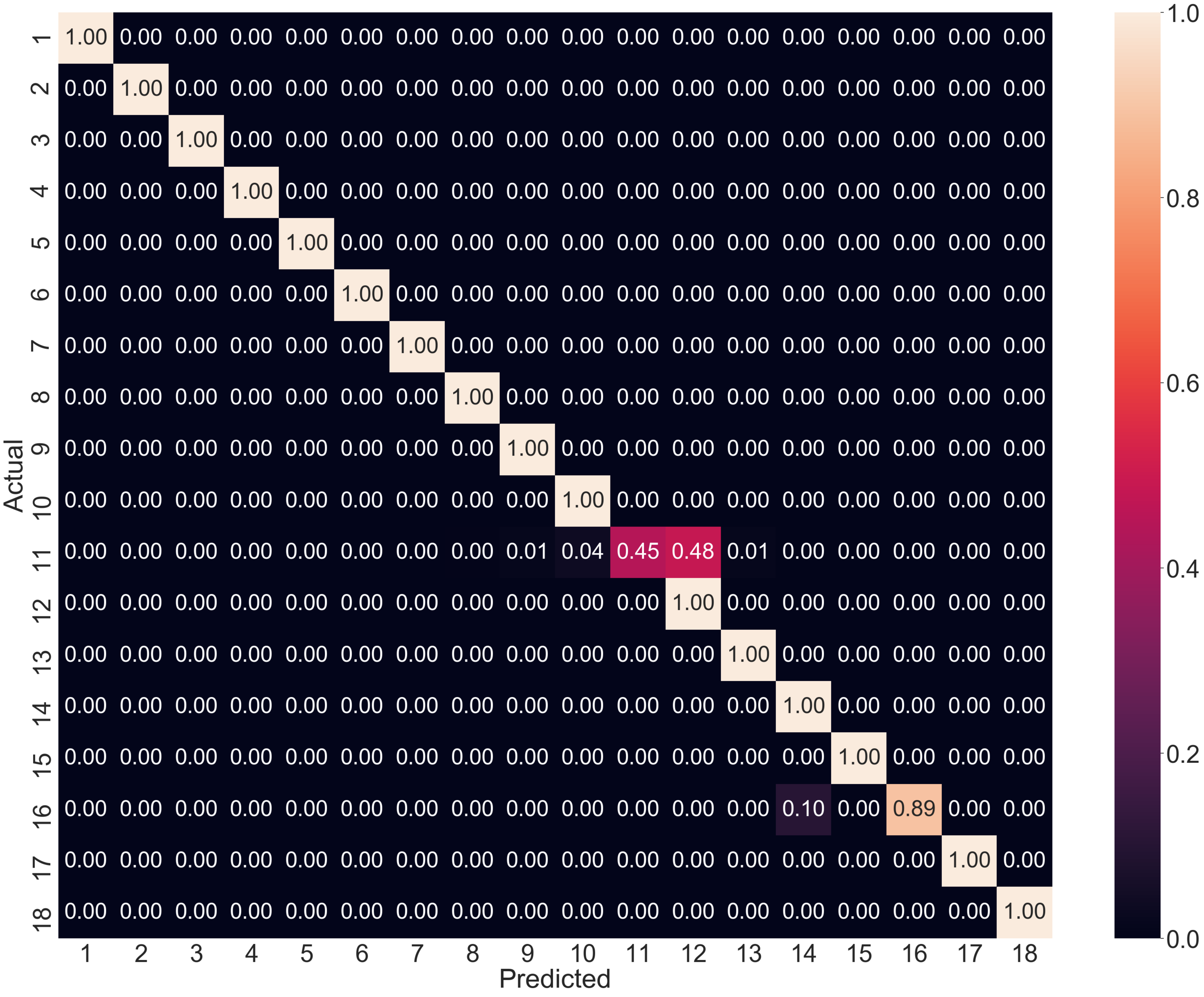}}\label{fig:EVAL:agent_cm}}
	\caption{Evaluation of $Agent_5$. The illustration shows the classification results of $M_{pre}(S_{robot})$ for the best poses within a sequence of 10 steps achieved by the agent. The results corresponds to the output of Table \ref{tab:result}.}
	\label{fig:EVAL:agent}
\end{figure}

Choosing $Agent_5$, we evaluate a sequence of actions of 1000 random starting poses per object on $S_{robot}$.
This is considered the baseline of our approach and results in an average classification confidence of 53\%.
The result in Table \ref{tab:result} shows an average improvement of 11\% for the first selected pose and an average improvement of 19\% for the best pose within the sequence.
This results in an absolute average of 64\% and 72\%, respectively, which shows that we find better viewpoints within the sequence.
Nevertheless, we cannot see any correlation between the performance of model $M_{pre}(S_{robot})$ and the improvements in this case either.
Empirically, we have observed that different agents achieve almost similar total results, but may differ in the improvement of individual objects.
The best pose is achieved within 4.28 steps on average.
While some objects reach the best pose in the first viewpoints, $O_{10}$ and $O_{18}$ show the best pose in the final one, which means that further steps could increase the confidence.
We see improvements for the first poses up to 71\% for some objects, although we have a negative change for individual others.
These values increase over time and usually end in a positive final result, with an improvement of up to 75\%.
This is illustrated as confusion matrix in Fig. \ref{fig:EVAL:agent_cm}, where we reach an average accuracy of 96.33\% compared to the initial 55.11\% accuracy (see Table \ref{tab:OC:prediction}).
Only the classification for $O_{11}$ and $O_{16}$ is not fully given, but the performance for $O_{11}$ can be increased by 10\%. 
\\
The final corresponding embedding in Fig. \ref{fig:EVAL:agent_umap} shows that most objects can be clustered well.
However due to the complexity of the block-constructed objects and therefore their obvious similarity, the embedding still spreads depending on the viewpoint of the object.
This becomes clear when looking at the individual outliers which, represent only a minority compared to the total of 1000 starting poses per object.

\begin{table}
		\vspace{0.06in}
	\renewcommand{\arraystretch}{1.05}
	\centering
	\caption{Evaluation of $Agent_5$. For each object $O_k$, a sequence of 10 actions has been evaluated 1000 times with random initial start poses used as our baseline. The table shows the relative change and the absolute values of the confidence difference for the first pose and best pose within a sequence of 10 steps. In addition, it shows the number of steps needed to reach this best pose.}
	\begin{tabular}{c|c|c|c|c|c|c}
		\hline
		\multirow{3}{*}{\begin{tabular}[c]{@{}l@{}}\textbf{Obj.}\\ $O_k$\end{tabular}}   
		& \multirow{3}{*}{\begin{tabular}[c]{@{}l@{}}\textbf{Start}\\ \textbf{Pose}\end{tabular}}
		         & \multicolumn{5}{c}{\textbf{Confidence $C(O_k)$}}                                                 \\ \cline{3-7} &
		         
		& \multicolumn{2}{c|}{First pose}                 & \multicolumn{3}{c}{Best pose}                  \\ \cline{3-7} 
		& & Abs.               & Change            & Abs.                & Change     & steps          \\ \hline
		1                     & 1.00                     & 1.00                   & 0.00                   & 1.00                   & 0.00    & 1              \\ \hline
		2                     & 0.61                     & 0.92                   & 0.31                   & 1.00                   & 0.39    & 3              \\ \hline
		3                     & 0.95                     & 0.99                   & 0.04                   & 0.99                   & 0.04    & 4              \\ \hline
		4                     & 1.00                     & 0.25                   & -0.75                  & 1.00                   & 0.00    & 2              \\ \hline
		5                     & 0.45                     & 0.68                   & 0.24                   & 0.86                   & 0.42    & 5             \\ \hline
		6                     & 0.51                     & 0.79                   & 0.29                   & 0.79                   & 0.29    & 1              \\ \hline
		7                     & 0.53                     & 0.64                   & 0.11                   & 0.65                   & 0.12    & 7              \\ \hline
		8                     & 0.18                     & 0.75                   & 0.57                   & 0.75                   & 0.57    & 1              \\ \hline
		9                     & 0.05                      & 0.74                   & 0.69                   & 0.80                   & 0.75    & 6              \\ \hline
		10                    & 0.25                     & 0.97                   & 0.71                   & 0.98                   & 0.71    & 10              \\ \hline
		11                    & 0.56                     & 0.31                   & -0.24                  & 0.31                   & -0.24   & 1              \\ \hline
		12                    & 0.39                     & 0.58                   & 0.18                   & 0.58                   & 0.18    & 1            \\ \hline
		13                    & 0.14                     & 0.30                   & 0.16                   & 0.43                   & 0.29    & 9             \\ \hline
		14                    & 0.54                      & 0.72                   & 0.19                   & 0.80                   & 0.26    & 3              \\ \hline
		15                    & 0.72                     & 0.42                   & -0.30                  & 0.52                   & -0.19   & 2              \\ \hline
		16                    & 0.52                     & 0.19                   & -0.33                  & 0.37                   & -0.16   &  8             \\ \hline
		17                    & 0.91                     & 0.67                   & -0.24                  & 0.67                   & -0.24   & 3            \\ \hline
		18                    & 0.21                     & 0.51                   & 0.30                   & 0.52                   & 0.31    & 10              \\ \hline
		\multicolumn{1}{c|}{\textit{\textbf{avg}}}& \textit{\textbf{0.53}} & \textit{\textbf{0.64}} & \textit{\textbf{0.11}} & \textit{\textbf{0.72}} & \textit{\textbf{0.19}} & \textit{\textbf{4.28}} \\ \hline
	\end{tabular}
	\label{tab:result}
\end{table}

\section{Conclusion}
\label{sec:conclusion}
In this work, we have presented a novel approach to find a sequence of next-best-views to classify objects moved by a robot arm.
Our iterative process between object classification and next-best-view estimation using SAC as a deep reinforcement learning method can determine the next-best poses of our target object to significantly increase the classification performance.
Moreover, we showed a significant improvement in object classification for the TEOS dataset, but it must be taken into account that we only used a reduced variant, halving the number of objects by excluding the distractors.
The dataset shows the non-trivial object classification, due to the self-occlusion of the objects.
Since we already have to work with the occlusion of the robot arm relative to the objects, we recognized problems with the classification of objects in the robot arm.
We showed that by using pre-trained models of simpler complexity we achieve better results for the more complex models of both, object classifier and next-best-view estimator.
Additionally, the use of the confidence difference as classification and reward metric instead of the usual confidence contributes to a better representation of the prediction distribution of the dataset during classification.
\\
We further hope to investigate the impact of the robot arm occlusion in addition to existing object self-occlusion to progress the next-best-view research for Active Object Recognition.
Future work would include the full $L_1$ TEOS dataset \cite{solbach2021blocks} and extend the simulation to include sensor-noise, depth of field, and shadowing.
In addition, a grasping approach could be used to grasp objects at the elements where the average occlusion to the robot arm is the lowest.



\section*{APPENDIX}
\begin{table}[H]
\caption{SAC Hyperparameters}
\renewcommand{\arraystretch}{1.1}
\centering
	\begin{tabular}{l|l}
		\hline
		\bfseries Parameter & \bfseries Value \\
		\hline
		batch size & 256 \\
		tau & 0.005 \\
		entropy coefficient & auto \\
		target update interval & 1 \\
		gradient steps & 1 \\
		target entropy & auto \\
		action noise & None \\
	    random exploration & 0 \\
		batch size & 256 \\
		tau & 0.005 \\
		entropy coefficient & auto \\
		target update interval & 1 \\
		gradient steps & 1 \\
		target entropy & auto \\
		action noise & None \\
		random exploration & 0 \\
		\hline
	\end{tabular}
	\label{tab:SAC:params}
\end{table}

\label{06_appendix}

\section*{ACKNOWLEDGMENT}



\end{document}